\documentclass[runningheads]{llncs}

\usepackage{eccv}

\usepackage{eccvabbrv}
\usepackage{amssymb}

\usepackage{graphicx}
\usepackage{booktabs}

\usepackage{xcolor}

\newcommand{\new}[1]{\textcolor{black}{#1}}

\usepackage[accsupp]{axessibility}  %

\usepackage{hyperref}
\usepackage{multirow}

\usepackage{orcidlink}

\newcommand\blfootnote[1]{%
  \begingroup
  \renewcommand\thefootnote{}%
  \NoHyper\footnote{#1}\endNoHyper
  \addtocounter{footnote}{-1}%
  \endgroup
}

\begin{document}

\title{Joint Flow Matching Enables Continuous Dose-Conditioned Cell Morphing}

\author{Lea Bogensperger\inst{1}\orcidlink{0009-0005-5765-056X} \and
Manuela Merlo\inst{1}\orcidlink{0009-0001-5814-3720} \and
Martin Baumgartner\inst{2}\orcidlink{0000-0001-9539-7204} \and
Michael Krauthammer\inst{1}\orcidlink{0000-0002-4808-1845} \and
Bernard Ciraulo\inst{2}\orcidlink{0000-0002-7191-8594}}

\authorrunning{L.~Bogensperger et al.}

\institute{University of Zurich
\email{\{lea.bogensperger,manuela.merlo,michael.krauthammer\}@uzh.ch}\\
\and
University Children’s Hospital Zurich\\
\email{\{martin.baumgartner,bernard.ciraulo\}@kispi.uzh.ch}}

\maketitle

\begin{abstract}
Generative modeling has shown increasing promise for predicting cellular perturbation effects under chemical compound treatments. Existing approaches either model perturbation as a distribution-to-distribution mapping without explicit concentration handling, or treat concentration as a discrete class label, precluding continuous dose control. 
We introduce a joint flow matching approach that simultaneously models cell latents and drug concentration via a dual-timestep formulation, enabling dose-conditioned single-cell morphing through the invertibility of flow matching. The joint formulation induces a monotonic dose-response geometry in latent space and additionally supports concentration estimation from cell morphology. As proof of concept, we further demonstrate generalization to an unseen dose held out during training. Empirically, our method achieves competitive or improved per-concentration metrics on two compounds compared with representative baselines, while enabling capabilities structurally unavailable to discrete-class methods.
  \keywords{Flow Matching \and Cell Perturbation \and Dose-Response 
Modeling \and Generative Modeling \and Cellular Morphology \and Fluorescence Microscopy }
\end{abstract}

\section{Introduction}
\label{sec:intro}

In cellular biology, morphological features often reflect cellular state and function \cite{wu2020}.
Optical microscopy is the dominant technique for capturing cellular morphology, commonly through fluorescence labeling and imaging of the protein or cellular compartment of interest \cite{hickey2022}. This approach provides rich information on cellular responses to external chemical perturbations.
Image-based response characterization is particularly relevant in drug development and drug-response profiling in oncology, where large panels of drugs are screened across a range of concentrations to score compounds or treatment efficacy \cite{way2021}. \blfootnote{\texttt{https://github.com/leabogensperger/joint-flow-matching-cells/}.}%
Such assays have historically relied on hand-designed morphological features~\cite{bray2016,caicedo2022}.
The scale of these experiments has grown substantially with modern automated microscopes, which enable extensive sampling of cellular populations across vast parameter spaces, generating large image collections and, with them, statistically robust characterization of morphological variation \cite{chandrasekaran2024}.
However, cellular populations are typically heterogeneous, and treatments often differ only subtly—whether due to similar chemical composition or fine-grained concentration screening. As a result, the morphologies observed in these experimental settings span a continuous space rather than falling into discrete categories, making assessment through classical computer vision approaches challenging \cite{serrano2026}.
An approach capable of modeling this continuum of morphologies and 
quantifying dose-dependent morphological drift would enable estimation 
of minimal effective doses and comparison of closely related compounds, 
offering meaningful benefits for compound screening.

Recent generative models offer a promising route toward this goal, 
learning to predict cellular perturbation responses directly from image 
data~\cite{zhang2025cellflux,zhang2026cellfluxv2,bourou2024phendiff,palma2025predicting}.
CellFlux~\cite{zhang2025cellflux} and its latent-space extension CellFluxV2~\cite{zhang2026cellfluxv2} frame the problem as a distribution-to-distribution transformation from control cells to treated cells using flow matching~\cite{lipman2022flow,liu2022flow}, but do not include concentration as a variable in their formulation. IMPA~\cite{palma2025predicting} learns compound-level style codes via a Generative Adversarial Network (GAN)~\cite{goodfellow2020generative}, again without an explicit concentration representation. PhenDiff~\cite{bourou2024phendiff} does condition on concentration using conditional diffusion models~\cite{ho2020denoising}, but treats it as a categorical class label, which prevents interpolation to 
unseen doses and does not capture the continuous nature of 
dose-response effects. None of these methods jointly models the distribution over cell images and concentration.

\begin{figure}[htbp]
    \centering
    \includegraphics[width=\linewidth]{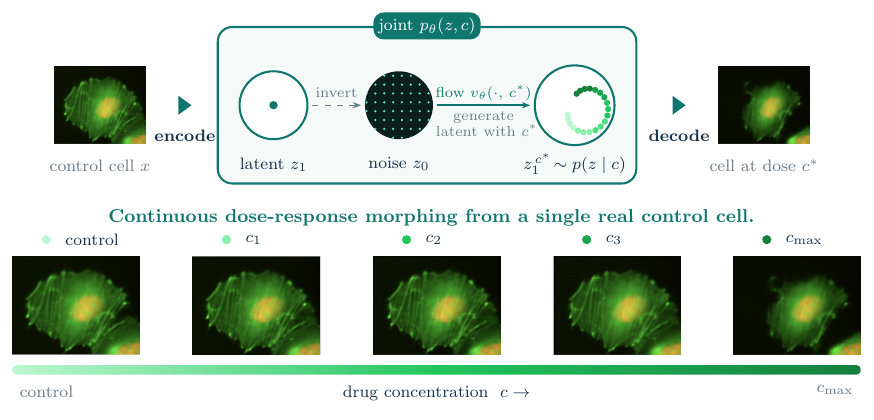}
    \caption{\textbf{Joint Flow Matching for continuous dose-response cell 
    image generation.} 
    Given a real control cell, our model encodes it into a VAE 
    latent $z_1$, inverts the flow to noise $z_0$, and re-integrates the forward ordinary differential equation (ODE) at target concentrations $c^\star$ to generate morphed cells 
    across a continuous dose range. The dual-timestep formulation 
    jointly models $p(z, c)$, enabling continuous concentration control.}
    \label{fig:teaser}
\end{figure}

We introduce a joint flow matching approach for continuous dose-response cell image generation over latent representations of cell images and drug concentration (Fig.~\ref{fig:teaser}). Our contributions are as follows:
\begin{itemize}
    \item We extend latent flow matching to jointly model the 
    distribution $p(z, c)$ over cell latents and drug concentrations 
    via a dual-timestep formulation.
    \item Our method enables dose-conditioned cell morphing by 
    leveraging the invertibility of flow matching, matching or 
    exceeding concentration-aware baselines on per-concentration 
    metrics while remaining competitive on concentration-agnostic evaluations.
    \item We show that joint modeling induces a monotonic mean 
    dose-response trajectory in latent space and improves concentration-specific fidelity over discrete-class baselines.
    \item As proof of concept, we hold out one concentration during 
    training and show generation at that unseen dose, a capability 
    structurally unavailable to discrete-class methods.
\end{itemize}

\section{Related Work}
\label{sec:related_work}

\paragraph{Flow Matching and Latent Generative Models.}
Diffusion models~\cite{song2019generative,song2020score,ho2020denoising} and flow matching~\cite{lipman2022flow, liu2022flow} have become state-of-the-art for high-fidelity generative modeling. Latent-space extensions such as Stable Diffusion~\cite{rombach2022high, esser2024scaling} exploit variational autoencoder (VAE) compression to enable efficient high-resolution generation while maintaining or improving sample quality. Flow matching has also proven effective in medical and cellular imaging domains~\cite{bourou2024phendiff,bogensperger2025flowsdf, 
zhang2025cellflux}. Conditional generation is naturally supported within the flow matching framework by conditioning the velocity network on auxiliary information, which can further be enhanced at inference via classifier-free guidance (CFG)~\cite{ho2022classifier}. Recent joint diffusion and flow approaches~\cite{bao2023one, li2025omniflow} extend single-modality generation to multiple modalities by introducing a separate timestep per modality.
Finally, invertibility is a key property of flow matching that enables image editing through inversion in noise space~\cite{hertz2022prompt, rout2024semantic}; recent work also explores inversion-free variants~\cite{kulikov2025flowedit}. 

\paragraph{Generative Models of Cellular Perturbation Response.}
Recent generative models targeting drug-induced morphological changes adopt a variety of architectural and methodological formulation choices. 
Because paired untreated–treated observations are typically unavailable, 
generative approaches model control and treated cells as distributions 
rather than individual correspondences. Flow matching enables CellFlux~\cite{zhang2025cellflux} to learn a continuous transformation between these distributions without paired examples. However, without additional constraints, aligning distributions in aggregate may miss subtle biological effects: the model captures broad distributional shifts rather than the specific treatment-induced changes.
Its extension CellFluxV2~\cite{zhang2026cellfluxv2} further moves the setup into a latent space and introduces a two-stage training procedure with noisy interpolants~\cite{albergo2025stochastic}. Neither variant models concentration as an explicit variable in their formulation. IMPA~\cite{palma2025predicting} and PhenExplain~\cite{lamiable2023revealing} use style-based GANs~\cite{goodfellow2020generative,karras2019style} to translate control cells to treated ones via per-perturbation style codes, but neither incorporates concentration in their formulation. Closest to our setting, PhenDiff~\cite{bourou2024phendiff} employs a class-conditional diffusion model~\cite{ho2020denoising,song2020denoising} with image inversion to translate cells between phenotypes, but treats concentration as a discrete class label, preventing continuous dose interpolation.

Our method combines latent flow matching (CellFluxV2) with 
inversion-based editing (PhenDiff), but unlike prior work that treats 
concentration as a discrete label or ignores it, we jointly model the 
(image, concentration) distribution with a continuous scalar. This enables source-conditioned cell morphing across doses and concentration estimation in a single framework.

\section{Method}
\label{sec:method}

Let ${x} \in \mathbb{R}^{C \times H \times W}$ denote a cell image with 
$C=3$ channels and spatial resolution $H \times W = 256 \times 256$, 
associated with a drug concentration $c \in \mathbb{R}_{\geq 0}$ (in $\mu$M). 
A pretrained VAE~\cite{kingma2013auto} with encoder $\mathcal{E}$ and decoder $\mathcal{D}$ maps images to a latent representation ${z} = \mathcal{E}({x}) \in \mathbb{R}^{d_z \times h \times w}$ with $d_z = 4$ latent channels and spatial resolution $h \times w = 32 \times 32$, yielding a spatial compression factor of $8\times$.

\subsection{Latent-Space Embedding}
\label{sec:vae}
\new{Similar to Latent CellFlux~\cite{zhang2026cellfluxv2}, we operate in a compressed latent space to enable efficient generative modeling. We therefore train} 
a VAE with encoder 
$\mathcal{E}_\psi: \mathbb{R}^{C \times H \times W} \to \mathbb{R}^{d_z \times h \times w}$ 
and decoder 
$\mathcal{D}_\psi: \mathbb{R}^{d_z \times h \times w} \to \mathbb{R}^{C \times H \times W}$, 
parameterized by $\psi$. This follows the latent diffusion 
paradigm~\cite{rombach2022high}: the VAE compresses images to a semantic 
latent space where the generative model (Sec.~\ref{sec:flow_matching}) 
operates efficiently, and the VAE is frozen for all downstream training. 
The VAE is trained by minimizing a combination of a structural 
similarity (SSIM) loss with an $\ell_1$ pixel error and a 
Kullback--Leibler divergence aligning the encoder posterior 
$q_\psi(z \mid x) = \mathcal{N}(\mu_\psi(x), \sigma_\psi^2(x))$ with a 
standard normal prior $p(z) = \mathcal{N}(0, I)$:
\begin{equation}
    \mathcal{L}_{\text{VAE}}(\psi; x) =
    \alpha \, \mathcal{L}_{\text{SSIM}}(\hat{x}, x)
    + (1 - \alpha) \, \mathcal{L}_{\text{L1}}(\hat{x}, x)
    + \beta \, D_{\text{KL}}\!\left(q_\psi(z \mid x) \,\|\, p(z)\right),
    \label{eq:vae_loss}
\end{equation}
where $\alpha, \beta \in \mathbb{R}^+$ weight the individual loss 
components, $\hat{x} = \mathcal{D}_\psi(\mathcal{E}_\psi(x))$ is the 
reconstruction, and $\mathcal{L}_{\text{SSIM}}(\hat{x}, x) = 1 - \operatorname{SSIM}(\hat{x}, x)$. The VAE itself is concentration-agnostic, concentration is only used downstream by the flow model (Sec.~\ref{sec:flow_matching}).

\subsection{Joint Flow Matching for Dose-Conditioned Generation}
\label{sec:flow_matching}

We train one flow model per drug over the joint distribution 
$p(z, {c})$ of latent images and their corresponding concentration.
\new{Similar to PhenDiff~\cite{bourou2024phendiff} which is based on DDIM inversion, we use the invertibility of flow matching. However, we model the concentration as a continuous scalar instead of discrete classes, and further use a joint dual timestep formulation to learn the joint distribution of latent images and their concentrations, which has been successfully done for other multi-modal settings~\cite{li2025omniflow,bao2023one}.}

\paragraph{Flow Matching Background.}
Flow matching~\cite{lipman2022flow, liu2022flow} learns a velocity 
field $v_\theta$ that transports samples from a simple prior 
$z_0 \sim \mathcal{N}(0, I)$ to the data distribution 
$z_1 \sim p_{\text{data}}$. {Training considers the linear interpolation:
\begin{equation}
  z_t = (1-t) z_0 + t z_1,  \;\; t \in [0, 1],
\end{equation}
whose velocity is constant and given by $z_1 - z_0$.
The conditional flow matching objective regresses the predicted velocity $ v_\theta(z_t, t)$ toward this target:}
\begin{equation}
    \mathcal{L}_{\text{CFM}} = 
    {\mathbb{E}_{z_0, z_1,t}}
    \left\| v_\theta(z_t, t) - (z_1 - z_0) \right\|_2^2.
\end{equation}

\noindent Sampling discretizes the ordinary differential equation (ODE) $dz_t/dt = v_\theta(z_t, t)$ into $K$ Euler steps of size $\Delta t = 1/K$: $z^{(k+1)} = z^{(k)} + \Delta t \, v_\theta(z^{(k)}, t^{(k)})$ with $t^{(k)} = k/K$, starting from $z^{(0)} = z_0$.

\paragraph{Invertibility.}
Because sampling is a deterministic ODE, the map from prior to data is invertible: integrating $v_\theta$ backwards in time from $z_1$ recovers $z_0$:
\begin{equation}
    z_0 = z_1 - \int_0^1 v_\theta(z_t, t)\, dt.
\end{equation}
This holds provided $v_\theta(\cdot, t, c)$ is sufficiently regular in $z$ and no noise is injected at inference. %
In practice we invert with Euler steps evaluated at the known iterate:
\begin{equation}
    z^{(k-1)} = z^{(k)} - \Delta t\, v_\theta\!\left(z^{(k)}, t^{(k)}\right).
\end{equation}

\paragraph{Joint Formulation with Dual Timesteps.}

{To model the joint distribution $p(z, c)$, we introduce 
\emph{two independent timesteps}, $t_z$ for the latent and $t_c$ for the concentration, inspired by previous works on multi-modal generative learning~\cite{bao2023one,li2025omniflow}. The timesteps are sampled uniformly $t_z, t_c \sim \mathcal{U}(0,1)$.}

{The latent variable \(z\) and the concentration \(c\) are then noised independently according to their respective timesteps:}
\begin{align}
    z_{t_z} &= (1 - t_z)\, z_0 + t_z\, z_1, 
    &z_0 &\sim \mathcal{N}(0, I),\; z_1 = \mathcal{E}(x), \\
    c_{t_c} &= (1 - t_c)\, c_0 + t_c\, c_1,
    &c_0 &\sim \mathcal{N}(0, 1),\; c_1 =c.
\end{align}
The network predicts velocities for both modalities with $v_z \in \mathbb{R}^{d_z \times h \times w}$ and $v_c \in \mathbb{R}$:
\begin{equation}
    (v_z, v_c) = v_\theta(z_{t_z}, c_{t_c}, t_z, t_c),
\end{equation}
trained with the joint objective ($\lambda_c \in \mathbb{R}^+$ balances both terms):
\begin{equation}
    \mathcal{L}_{\text{CFM}}(\theta) = 
    \mathbb{E}_{(z_0, z_1), (c_0, c_1), t_z, t_c}\!\left[ 
    \|v_z - (z_1 - z_0)\|_2^2 
    + \lambda_c\, \|v_c - (c_1 - c_0)\|_2^2 
    \right].
\end{equation}

\paragraph{Sampling Modes.}
The two independent timesteps enable three distinct inference regimes. 
\textbf{Joint sampling} integrates both $t_z, t_c: 0 \to 1$ 
simultaneously to draw new $(z, c)$ pairs from the learned joint 
distribution; while beyond the scope of this work, this mode could 
support downstream tasks such as synthetic data augmentation. 
Our experiments primarily target \textbf{dose-conditioned image generation}, 
which fixes $t_c = 1$ (target concentration held constant) and 
integrates $t_z: 0 \to 1$ to sample $z \sim p(z \mid c)$, followed by 
decoding via $\mathcal{D}$; this corresponds to our dose-conditioned 
generation and inversion-based morphing experiments 
(Sec.~\ref{sec:morphing}). Conversely, \textbf{concentration inference} 
fixes $t_z = 1$ (image latent held constant) and integrates 
$t_c: 0 \to 1$ to sample $c \sim p(c \mid z)$, allowing the model to 
predict the concentration of a given cell.

\subsection{Inversion-Based Cell Morphing}
\label{sec:morphing}
The invertibility of flow matching enables cell morphing across drug 
concentrations, analogous to inversion-based editing with diffusion~\cite{hertz2022prompt} and its recent adoption for cellular perturbation via PhenDiff~\cite{bourou2024phendiff}.
Given a real control image $x$, we 
encode it as $z_1 = \mathcal{E}(x)$ and integrate the flow \emph{backward} 
at the control conditioning value $c=0$ using $K$ Euler steps 
with $\Delta t = 1/K$, with the concentration timestep fixed at $t_c = 1$, 
to recover the corresponding noise:
\begin{equation}
    z^{(k-1)} = z^{(k)} - \Delta t~v_z\!\left(z^{(k)}, c_{\text{control}}, 
    t_z{=}\tfrac{k}{K}, t_c{=}1\right), 
    \quad k = K, K{-}1, \dots, 1,
\end{equation}
starting from $z^{(K)} = z_1$ and ending at the inverted noise 
$z_0 := z^{(0)}$. Integrating the forward ODE from the same $z_0$ at a 
target concentration $c^\star$ then yields a morphed latent:
\begin{equation}
    z^{(k+1)} = z^{(k)} + \Delta t~v_z\!\left(z^{(k)}, c^\star, 
    t_z{=}\tfrac{k}{K}, t_c{=}1\right), 
    \quad k = 0, 1, \dots, K{-}1,
\end{equation}
yielding $z_1^{c^\star} := z^{(K)}$ which is decoded as $\hat{x}^{c^\star} = \mathcal{D}(z_1^{c^\star})$. Sharing the same $z_0$ across concentrations yields a family of latents that differ only through the concentration conditioning, producing a morph strip for a single source cell.

\section{Experiments}
\label{sec:experiments}
We evaluate our method on a microscopy imaging dataset of a pediatric brain tumor cell line, spanning a panel of two compounds (C7 and CK-666) at six concentrations each (five doses plus solvent controls). The VAE is trained jointly on all five compounds in the dataset, providing a shared latent space across drugs (see Sec.~\ref{sec:supp_ablation} for an ablation). Per-compound flow matching models are then trained on top of this shared latent space. 
We compare against three representative baselines from the literature: PhenDiff~\cite{bourou2024phendiff}, IMPA~\cite{palma2025predicting}, and Latent CellFlux (our reimplementation of CellFluxV2~\cite{zhang2026cellfluxv2}). 
Main-text experiments focus on C7; corresponding results for CK-666 are provided in the supplementary material.

\subsection{Dataset}
\label{sec:dataset}

\paragraph{Sample.} 
The pediatric brain tumor medulloblastoma cell line model ONS-76 was used to evaluate a panel of compounds selected for their ability to interfere with cell migration through modulation of cytoskeletal dynamics, a key biophysical mechanism involved in metastasis \cite{Raudenska2023}. This panel included C7, a small-molecule ligand of FRS2 proposed to disrupt the FRS2–FGFR interaction\cite{Kumar2023}; CK-666, an inhibitor of the Arp2/3 complex \cite{Hetrick2012}; famlasertib, a small-molecule inhibitor of the MAP4K family of Ser/Thr kinases \cite{Bos2019}\cite{Schonholzer2026}; para-nitroblebbistatin, an inhibitor of non-muscle myosin II ATPase activity  \cite{Kepiro2014}; Y-27632, a selective ROCK1/2 inhibitor \cite{Uehata1997}; and DMSO solvent as control.
To enable fluorescence live-cell imaging, ONS-76 cells were labeled by lentiviral transduction with LifeAct-EGFP and mCherry-Nuc9. LifeAct-EGFP was used to visualize F-actin, while mCherry-Nuc9 was used to mark the nucleus.

\paragraph{Microscopy.} 
Widefield fluorescence imaging of both channels was performed on a Nikon Ti2 microscope equipped with a 20× objective and GFP and Cy3 filter sets, using tiled acquisition of 6×6 fields of view per well with 10\% overlap (tile size 2424×2424 px, pixel size 0.33 µm/px, 16-bit depth); individual tiles were automatically stitched using Nikon acquisition software. Cells were seeded in imaging-grade glass-bottom 96-well plates, with a minimum of two wells replicates per drug and concentration tested.

\paragraph{Single-cell Dataset Creation.}
From the stitched tiled images, single-cell crops of 256×256 pixels centered on each detected cell were extracted to build an RGB single-cell image dataset. Cell detection was performed on 10-fold downscaled images using Cellpose (v4.1.1) with default parameters. For each extracted single-cell image, the green (F-actin) and red (nuclear) channel dynamic ranges were normalized to 8-bit prior to saving, while the blue channel was set to zero.

\paragraph{Overview.}
Our modeling setup involves two components: a single VAE trained jointly on the training splits of all five compounds in the dataset, providing a shared latent space; and per-compound flow-matching models trained on top of this latent space. Reported experiments target C7 and CK-666 \new{(see Fig.~\ref{fig:dataset} for some example images)}; the shared VAE is trained on all available compounds to encourage a general representation of cellular morphology (see Sec.~\ref{sec:supp_ablation} for an ablation).

\paragraph{Data Handling.}
Regions failing standard image quality checks (e.g., saturation, focus, signal-to-noise) were excluded. Each compound's images are split into train/validation/test partitions in a 70/15/15 ratio, stratified by concentration to preserve the dose distribution across splits. Image counts per compound (including shared DMSO controls) are: 25{,}384 for C7, 25{,}603 for CK-666, 23{,}007 for famlasertib, 25{,}068 for paranitro-blebbistatin, and 25{,}823 for Y-27632.
Since concentrations span multiple orders of magnitude, we use $\log_{10}$-transformed values; controls are assigned $c_{\text{control}} = \log_{10}(c_{\min}) - 0.5$, where $c_{\min}$ is the smallest non-zero training concentration.

\subsection{Settings}
\paragraph{Latent Encoder.}
Our VAE follows a standard convolutional encoder-decoder design with 
three downsampling stages (each halving the spatial resolution, base 
channels 64 doubled per stage) built from ResNet blocks, mapping 
$3 \times 256 \times 256$ images to $4 \times 32 \times 32$ latents. The VAE is trained on the training splits of all five drugs and all concentrations for 200 epochs with AdamW (learning rate $10^{-4}$, batch size 32, warmup ratio 0.05), using loss weights $\alpha = 0.6$ (SSIM) and $\beta = 10^{-7}$ (KL).

\paragraph{Flow Model.}
The velocity network $v_\theta$ is a U-Net with self-attention at the bottleneck, base dimension 64, and channel multipliers $(1, 2, 4, 8)$ across four downsampling stages. Both timesteps $t_z, t_c$ and the concentration scalar $c$ are embedded via small MLPs with nonlinear GELU activations; the two time embeddings are concatenated and combined with the concentration embedding to form the conditioning signal injected into each residual block; a small MLP head predicts $v_c$ from the U-Net bottleneck features. We train one model per drug (\emph{Ours joint}, 37.8M parameters) for 3000 epochs with AdamW (learning rate $10^{-4}$, batch size 100, EMA decay 0.999), using $\lambda_c = 0.5$ (chosen empirically). Concentrations are $\log_{10}$-transformed as described in Section~\ref{sec:dataset}, see Sec.~\ref{sec:supp_ablation} for an ablation. 
Inference uses $K = 200$ Euler steps for forward generation and 
$K = 200$ for inversion \new{(see Sec.~\ref{sec:sampler_ablations} for an ablation over samplers)}. For ablation, we additionally 
train a concentration-conditional variant (\emph{Ours cond.\ only}, 31.5M parameters) with the same U-Net architecture but without the second timestep $t_c$, using classifier-free guidance with dropout probability $p = 0.15$ during training and guidance weight $w = 2$ at inference. \new{Concentration conditioning in this variant is injected via Feature-wise Linear Modulation (FiLM)~\cite{perez2018film}, though initial experiments showed standard embedding mechanisms perform comparably.}

\paragraph{Baselines.}
We compare against three representative methods: Latent 
CellFlux~\cite{zhang2025cellflux} (31.4M parameters, concentration-agnostic 
latent flow matching), IMPA~\cite{palma2025predicting} (74.4M parameters, 
pixel-space per-perturbation GAN), and PhenDiff~\cite{bourou2024phendiff} 
(99.5M parameters, pixel-space DDIM with discrete concentration classes).
We use publicly available code from PhenDiff and IMPA, both of which operate in pixel space. For CellFluxV2, which operates in latent space, we reimplement the method in our own VAE latent space; we refer to this reimplementation as Latent CellFlux. We use $K=200$ sampling steps for all flow-based methods and PhenDiff-standard DDIM inference for the diffusion baseline.

\paragraph{Evaluation Metrics.}
For \emph{inversion fidelity}, we report peak signal-to-noise ratio (PSNR) and structural similarity (SSIM)~\cite{wang2004image} on the test set, comparing each real control image against its round-trip reconstruction obtained after inverting the encoded control cell followed by re-generating it using the forward ODE.
For \emph{distributional fidelity}, we compute Fréchet Inception Distance (FID)~\cite{heusel2017gans} and Kernel Inception Distance (KID)~\cite{binkowski2018demystifying} per concentration and pooled 
across all concentrations, \new{using Inception-v3 features extracted from all available real training images and an equal number of generated cells ($N_{\text{real}} = N_{\text{gen}} = N$; $N \approx 850\text{--}1,250$ per dose, $N > 5,000$ pooled).} Generated cells are obtained by inverting real control cells and forward-integrating at the target concentration, similar to PhenDiff~\cite{bourou2024phendiff}. \new{Because FID is subject to finite-sample bias~\cite{chong2020effectively}, we report $N$ explicitly and emphasize the unbiased KID estimator alongside FID for robust per-concentration evaluation.}

\subsection{Generative Evaluation}
\label{sec:gen_eval}
We evaluate our joint flow matching model along three dimensions: round-trip reconstruction fidelity of the inversion pathway, distributional fidelity of generated cells against real observations at each concentration, and morphing quality against baselines. We report the main results on C7 in this section; corresponding results for CK-666 are provided in Sec.~\ref{sec:supp_ck666}.

\paragraph{Inversion Fidelity.}
Since dose-conditioned morphing relies on inverting a real control cell 
into noise before re-integrating the forward ODE at a target concentration 
$c^*$, we first assess the fidelity of the inversion. 
Table~\ref{tab:inversion} reports round-trip PSNR and SSIM on 100 control 
test images for C7 (see Tab.~\ref{tab:inversion_ck666} for CK-666), 
computed by encoding each image to its latent $z_1$, inverting to noise 
$z_0$ at the control conditioning value, and re-integrating the forward 
ODE back to the input. The metrics are upper-bounded by the VAE's own 
reconstruction quality (last row), since part of the observed error 
comes from the VAE bottleneck rather than the flow inversion itself. 
Reconstruction quality saturates with increasing ODE steps; we adopt 
$K = 200$ steps for inversion in all subsequent experiments. Notably, 
SSIM nearly matches the VAE upper bound, indicating that structural 
content is well-preserved through the flow round-trip; the larger PSNR 
gap reflects small pixel-level noise introduced by the ODE 
discretization, to which pixel-wise metrics are more sensitive than the 
patch-based SSIM.

\begin{table}[tb]
\centering
\caption{Inversion fidelity for C7 on the test set ($N{=}100$ control cells).}
\label{tab:inversion}
\small
\begin{tabular}{lcc}
\toprule
ODE steps $K$ & PSNR $\uparrow$ & SSIM $\uparrow$ \\
\midrule
10  & $27.79_{\pm 2.9}$ & $0.866_{\pm 0.04}$ \\
25  & $31.93_{\pm 3.0}$ & $0.894_{\pm 0.04}$ \\
50  & $33.81_{\pm 3.0}$ & $0.902_{\pm 0.04}$ \\
100 & $34.50_{\pm 3.0}$ & $0.905_{\pm 0.05}$ \\
200 & $34.70_{\pm 3.0}$ & $0.906_{\pm 0.05}$ \\
500 & $34.77_{\pm 3.0}$ & $0.907_{\pm 0.05}$ \\
\midrule
VAE only & $38.73_{\pm 3.1}$ & $0.913_{\pm 0.05}$ \\
\bottomrule
\end{tabular}
\end{table}

\begin{figure}[ht]
    \centering
    \includegraphics[width=\linewidth]{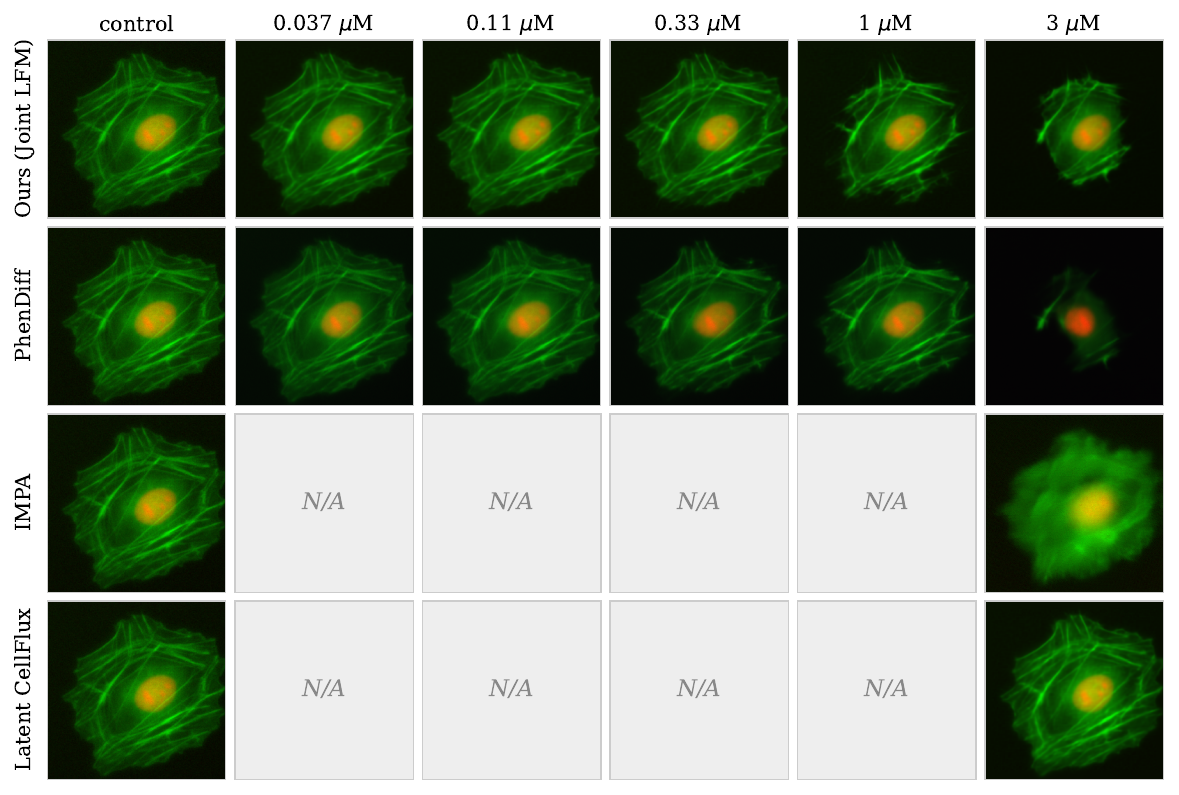}
    \caption{
        \textbf{Qualitative comparison of dose-response morphing for C7.}
        Each row shows a real control cell morphed across concentrations of C7 
        (columns: 0.037, 0.11, 0.33, 1.0, 3.0 $\mu$M). 
        Latent CellFlux and IMPA are shown only at control and morphed dose. Individual samples from concentration-agnostic methods can vary in visual quality across cells. %
        }
    \label{fig:qualitative_morphs}
\end{figure}

\paragraph{Qualitative Morphing.}
Figure~\ref{fig:qualitative_morphs} shows dose-conditioned morphs from our method and baselines that support dose-specific generation. Concentration-agnostic methods (IMPA, Latent CellFlux) formulate perturbation as a distribution-to-distribution transformation and are not designed to natively represent the subtle morphological changes characteristic of dose response.

\paragraph{Distribution Fidelity.}
Table~\ref{tab:c7_fid_results} reports FID and KID scores for C7 across five concentrations and pooled across the full dose range. To isolate the contribution of joint modeling, we include an ablation without the joint model and hence the second timestep $t_c$ (\emph{Ours cond.\ only}). Since Latent CellFlux and IMPA do not natively support dose-specific generation, as both model a single treated-cell distribution per compound, we report their metrics only pooled, as an upper reference on generation quality without concentration awareness. Among methods with explicit concentration modeling, our joint model achieves the best FID and KID at every concentration. While pixel-space IMPA achieves the lowest pooled FID overall (17.58), our joint model closes most of this gap despite operating in latent space.

\begin{table}[tb]
\centering
\caption{FID/KID comparison (lower is better) for C7. Per-concentration values 
shown for methods with explicit concentration modeling; concentration-agnostic 
baselines (Latent CellFlux, IMPA) are reported at pooled only, as they do not natively support concentration targeting. Best metrics per column are 
in \textbf{bold}, second-best \underline{underlined}.}
\label{tab:c7_fid_results}
\small
\begin{tabular}{llcccccc}
\toprule
& & \multicolumn{5}{c}{Concentration ($\mu$M)} & \\
\cmidrule(lr){3-7}
Method & Metric & 0.037 & 0.11 & 0.33 & 1.0 & 3.0 & Pooled\\[-0.4em]
 & & \tiny($N{=}892$) & \tiny($N{=}1028$) & \tiny($N{=}868$) & \tiny($N{=}1040$) & \tiny($N{=}1270$) & \tiny($N{=}5098$) \\
\midrule
Latent CellFlux~\cite{zhang2025cellflux}\textsuperscript{$\dagger$}
    & FID & --- & --- & --- & --- & --- & 25.19 \\
    & KID & --- & --- & --- & --- & --- & 1.90 \\
\midrule
IMPA~\cite{palma2025predicting}\textsuperscript{$\dagger$}
    & FID & --- & --- & --- & --- & --- & \textbf{17.58} \\
    & KID & --- & --- & --- & --- & --- & \textbf{0.97} \\
\midrule
PhenDiff~\cite{bourou2024phendiff}
    & FID & \underline{30.03} & \underline{29.74} & \underline{33.22} & \underline{31.90} & \underline{34.16} & 21.53 \\
    & KID & \underline{1.61}  & 1.73              & \underline{1.73}  & \underline{1.91}  & \underline{2.10}  & 1.68  \\
\midrule
Ours (cond.\ only) 
    & FID & 36.19 & 30.86 & 42.63 & 37.43 & 38.23 & 26.57 \\
    & KID & 2.25  & \underline{1.72}  & 2.93  & 2.45  & 2.65  & 2.10  \\
\midrule
\textbf{Ours (Joint Latent FM)}
    & FID & \textbf{30.02} & \textbf{27.20} & \textbf{27.50} & \textbf{27.85} & \textbf{25.02} & \underline{18.98} \\
    & KID & \textbf{1.58}  & \textbf{1.40}  & \textbf{1.37}  & \textbf{1.39}  & \textbf{1.28}  & \underline{1.25}  \\
\bottomrule
\end{tabular}
\vspace{2pt}
\begin{flushleft}
\footnotesize
\textsuperscript{$\dagger$} Latent CellFlux and IMPA do not natively support concentration conditioning; therefore we report their pooled results without concentration awareness.
\end{flushleft}
\end{table}

\subsection{Concentration Controllability}  
\label{sec:controllability}
While distributional metrics such as FID and KID capture overall generation quality, they do not directly assess whether generation is controllable across concentrations. We evaluate controllability along three dimensions: latent-space trajectory, cross-concentration specificity, and generalization to unseen doses using a holdout experiment.

\begin{figure}[tb]
\centering
\includegraphics[width=\linewidth]{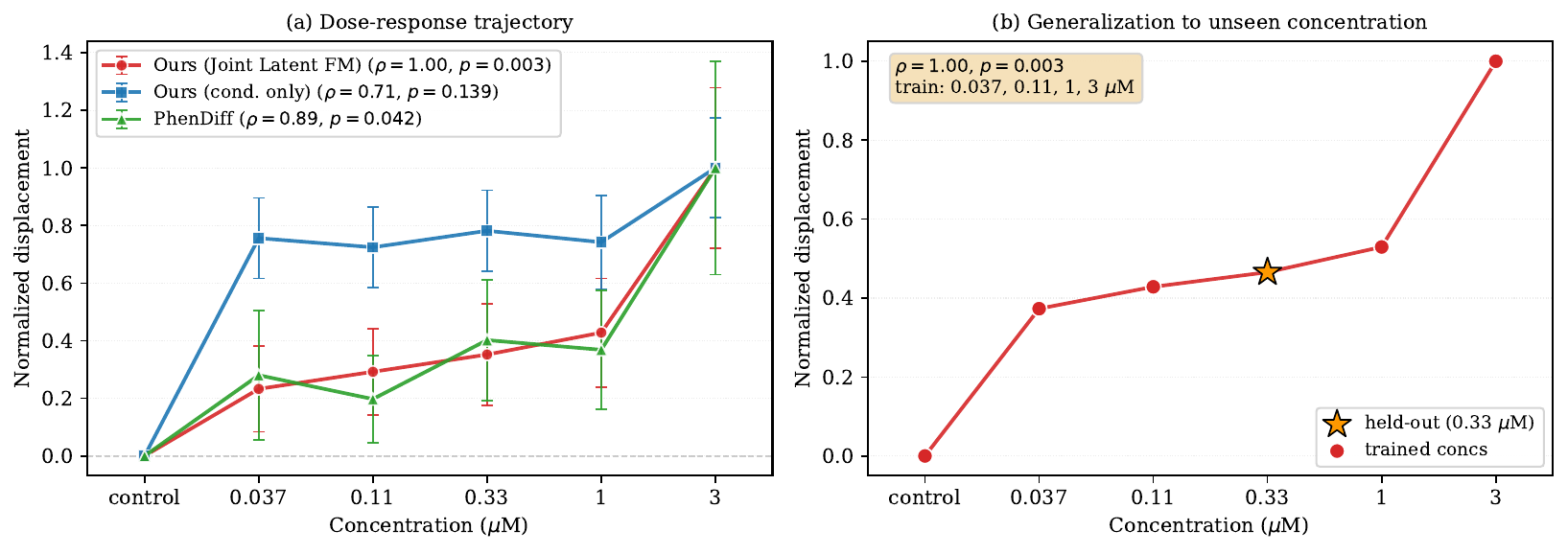}
\caption{
\textbf{Dose-response geometry and generalization to unseen concentrations for C7.}
Left: Normalized latent displacement from the reference as a function of drug concentration, comparing our joint model, our conditional-only ablation, and PhenDiff. Right: After retraining our joint model with $0.33\,\mu$M held out during training (train concentrations: $0.037, 0.11, 1.0, 3.0\,\mu$M), forward integration at the held-out dose (highlighted star) yields a monotonic position between its neighbors.}
\label{fig:trajectory}
\end{figure}

\paragraph{Latent Space Dose-Response.}
We sample a fixed noise vector $z^{(0)}$, integrate the forward ODE at each concentration, and measure latent displacement from the control trajectory.
Figure~\ref{fig:trajectory} shows the resulting curves and Spearman correlations ($n=100$ cells).
Our joint model produces a smooth, monotonic dose response ($\rho = 1.00$, $p = 0.003$), while removing the joint component or modeling concentration using discrete classes degrades this monotonic behaviour ($\rho = 0.71$, $p = 0.139$ and $\rho = 0.89$, $p = 0.042$, respectively). \new{Evaluating individual single-cell trajectories ($n=100$ cells) confirms strong per-cell monotonicity for our joint model with a median $\rho = 0.90$ (IQR: $0.70\text{--}1.00$), outperforming PhenDiff (median $\rho = 0.80$, IQR: $0.50\text{--}0.90$) and the conditioning-only baseline (median $\rho = 0.40$, IQR: $0.30\text{--}0.70$; Supplementary Fig.~\ref{fig:violin}).} %

\begin{figure}[tb]
\centering
    \includegraphics[width=\linewidth]{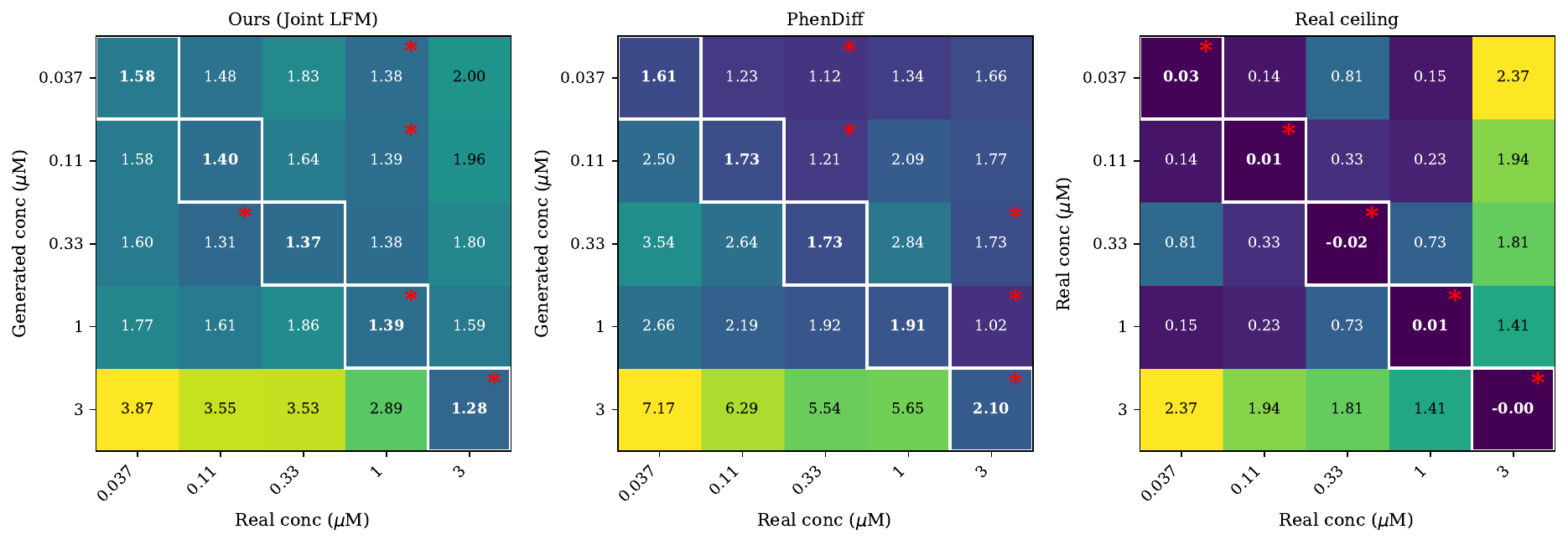}
\caption{
\textbf{Concentration specificity.}
KID between generated cells conditioned on concentration $c_i$ (rows) and real cells at concentration $c_j$ (columns). Diagonals are outlined; red asterisks mark each row's argmin. Panels (left to right): our joint model, PhenDiff, and the real-vs-real reference ceiling.}
\label{fig:conc_matrix}
\end{figure}

\paragraph{Concentration Specificity.}
While the trajectory analysis probes latent geometry, we further 
evaluate whether this structure translates into controllable generation at each concentration. For each pair $(c_i, c_j)$, we compute the KID between the distribution of generated cells conditioned on $c_i$ and the distribution of real cells at $c_j$; a controllable model produces diagonal-dominant matrices where the minimum along each row lies on the diagonal. 
Results are shown in Fig.~\ref{fig:conc_matrix}, including a real-vs-real ceiling. The ceiling shows the task's inherent difficulty: cells at $3\,\mu$M are well-separated from all doses, but low-dose distributions ($0.037$--$0.33\,\mu$M) morphologically overlap even in real data, upper-bounding achievable specificity. Nonetheless, our joint model achieves lower diagonal KID values than PhenDiff at every concentration, indicating closer distributional matching.

\paragraph{Generalization to Unseen Concentrations.}
To test whether continuous concentration modeling enables interpolation to unseen concentrations, we retrain our joint model on C7 with the middle concentration ($0.33\,\mu$M) held out (train concentrations: $0.037, 0.11, 1.0, 3.0\,\mu$M). At inference, forward integration at the held-out dose lands at the correct monotonic position between its neighbors in latent space (Fig.~\ref{fig:trajectory}b, orange star), achieving $\text{FID} = 33.66$ against real $0.33\,\mu$M cells --- only $\sim\!19\%$ above the mean FID at trained concentrations of the same model ($28.20$). Notably, this out-of-distribution result almost matches PhenDiff ($\text{FID} = 33.22$, Tab.~\ref{tab:c7_fid_results}), a discrete-class baseline that \emph{was} trained on $0.33\,\mu$M. Such continuous-dose generalization is structurally unavailable to discrete-class methods. \new{See Fig.~\ref{fig:holdout} for more held-out concentrations ($0.11\,\mu$M and $1\,\mu$M) on C7.}

\paragraph{Concentration Inference.}
A distinctive capability of our joint formulation is sampling from 
$p(c \mid z)$ to predict the dose corresponding to a given cell 
image. Fixing $t_z = 1$ and integrating $t_c: 0 \to 1$ from a noise 
sample, we obtain a concentration estimate $\hat{c}$. On the test set 
($N=1115$ cells across all training doses), we achieve top-1 accuracy 
of $34.0\%$ and $68.0\%$ within $\pm 1$ neighboring dose 
($5$-class classification). Accuracy is highest at the extreme 
concentrations ($44\%$ at $0.037\,\mu$M, $55\%$ at $3\,\mu$M) and drops 
at intermediate doses, consistent with the overlapping morphological distributions of neighboring low concentrations shown in Fig.~\ref{fig:conc_matrix}. \new{For reference, a supervised ResNet-18 classifier trained on the same real data achieves $48\%$ top-1 and $74.8\%$ within $\pm 1$ neighboring dose, as detailed in Sec.~\ref{sec:supp_baseline}.}

\subsection{Ablations}
\label{sec:supp_ablation}
We perform two ablations on the joint model's design choices for C7.
Table~\ref{tab:vae_ablation} shows that training the VAE on all drugs (our default) yields a more general latent representation than training on C7 alone, improving FID and KID at every concentration.
Since drug concentrations span multiple orders of magnitude, we 
condition the flow model on $\log_{10}$-transformed values rather than raw concentrations; Tab.~\ref{tab:log_ablation} shows that this improves FID and KID at every concentration.

\begin{table}[tb]
\centering
\caption{VAE training-data ablation on C7. Best values in \textbf{bold}.}
\label{tab:vae_ablation}
\small
\begin{tabular}{llcccccc}
\toprule
& & \multicolumn{5}{c}{Concentration ($\mu$M)} & \\
\cmidrule(lr){3-7}
VAE training & Metric & 0.037 & 0.11 & 0.33 & 1.0 & 3.0 & Pooled \\
\midrule
C7-only 
    & FID & 35.26 & 33.39 & 34.37 & 33.89 & 31.54 & 24.84 \\
    & KID & 2.18  & 2.09  & 2.14  & 2.06  & 2.01  & 1.94  \\
\midrule
Multi-drug
    & FID & \textbf{30.02} & \textbf{27.20} & \textbf{27.50} & \textbf{27.85} & \textbf{25.02} & \textbf{18.98} \\
    & KID & \textbf{1.58}  & \textbf{1.40}  & \textbf{1.37}  & \textbf{1.39}  & \textbf{1.28}  & \textbf{1.25}  \\
\bottomrule
\end{tabular}
\end{table}

\begin{table}[tb]
\centering
\caption{Concentration-transform ablation on C7. Best values in \textbf{bold}.}
\label{tab:log_ablation}
\small
\begin{tabular}{llcccccc}
\toprule
& & \multicolumn{5}{c}{Concentration ($\mu$M)} & \\
\cmidrule(lr){3-7}
Conc.\ transform & Metric & 0.037 & 0.11 & 0.33 & 1.0 & 3.0 & Pooled \\
\midrule
Raw ($c$)
    & FID & 30.27 & 27.72 & 29.94 & 28.50 & 27.02 & 20.31 \\
    & KID & 1.61  & 1.46  & 1.67  & 1.46  & 1.48  & 1.35  \\
\midrule
$\log_{10}(c)$
    & FID & \textbf{30.02} & \textbf{27.20} & \textbf{27.50} & \textbf{27.85} & \textbf{25.02} & \textbf{18.98} \\
    & KID & \textbf{1.58}  & \textbf{1.40}  & \textbf{1.37}  & \textbf{1.39}  & \textbf{1.28}  & \textbf{1.25}  \\
\bottomrule
\end{tabular}
\end{table}

\section{Conclusion}
\label{sec:conclusion}

\paragraph{Conclusion.}
We introduced joint latent flow matching for continuous 
dose-conditioned cell morphing, enabling generation at unseen 
concentrations and estimation of dose from cell morphology. 
Our joint formulation matches or exceeds baselines on distributional 
fidelity across two compounds, while producing qualitatively faithful 
dose-response trajectories.

\paragraph{Limitations.}
Our reconstruction fidelity is currently bounded by the frozen VAE 
bottleneck (Sec.~\ref{sec:vae}), which imposes a ceiling on generation 
quality regardless of the flow model's capability. Furthermore, 
concentration inference from single cells (Sec.~\ref{sec:controllability}) 
achieves only moderate accuracy, reflecting the inherent phenotypic 
variability of morphology at fixed doses.

\paragraph{Future Work.}
Replacing the VAE with a stronger encoder-decoder (e.g., the 
architecture used in Stable Diffusion LDM~\cite{rombach2022high}) could 
substantially raise the reconstruction ceiling. A second 
direction is to extend our framework to a single flow model 
conditioned jointly on drug identity and concentration, modeling 
multiple compounds within a unified formulation, \new{e.g., by adopting the drug conditioning mechanism in IMPA~\cite{palma2025predicting}.} Moreover, our 
current per-compound models already share a common latent space via 
the multi-drug VAE, enabling comparative analyses of latent 
dose-response trajectories across drugs to group compounds by their 
morphological effect.
\new{Finally, downstream validation of generated morphs using a classifier model trained on real images would strengthen the practical utility of our approach for perturbation-based cell morphing.} %

\section*{Acknowledgements}
\new{L.B. and M.M. acknowledge funding by the University Research Priority
Program (URPP) Human Reproduction Reloaded of the
University of Zurich. L.B. is further supported by the Swiss National Science Foundation under grant 10003518. M.B. and B.C. were supported by the Swiss National Science Foundation Sinergia grant CRSII5\_202245/1.}

\bibliographystyle{splncs04}
\bibliography{main}

\clearpage
\appendix
\section{Supplementary Material}

\subsection{Dataset \& Qualitative Real Samples}

\begin{figure}[htb]
\centering
\includegraphics[width=\linewidth]{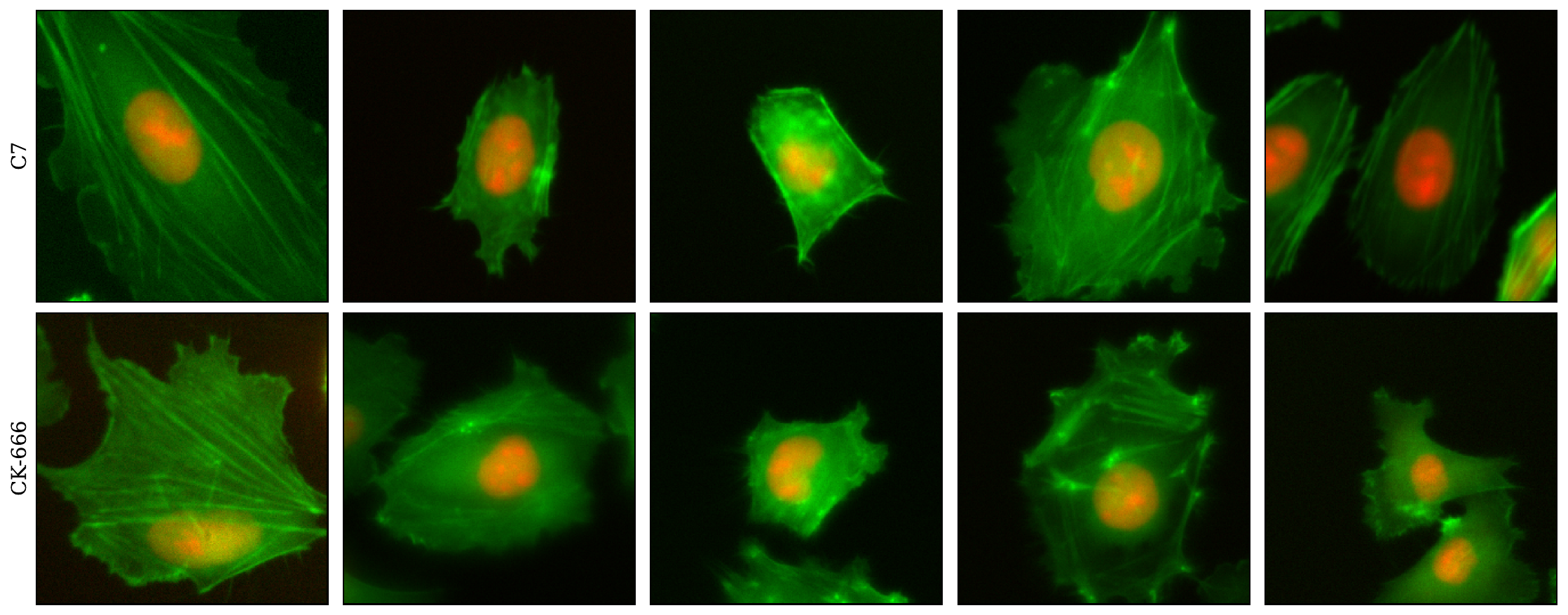}
\caption{\textbf{Qualitative training set samples for ONS76 cell line under compound treatments.} Representative real microscopy images of ONS76 cells treated with C7 (top row, 3$\mu$M) and CK-666 (bottom row, 10$\mu$M) from the training split.}
\label{fig:dataset}
\end{figure}

\new{Figure~\ref{fig:dataset} shows qualitative samples of the training split for the maximum concentrations of C7 (top row) and CK-666 (bottom row).}

\subsection{Experimental Results on CK-666}
\label{sec:supp_ck666}

To assess generalization beyond C7, we evaluate the generative capabilities of our of joint model on a second compound, CK-666. We show quantitative results for inversion fidelity and per-concentration distributional fidelity. 

\paragraph{Inversion Fidelity.}
Table~\ref{tab:inversion_ck666} reports the inversion fidelity for CK-666. Similarly as for C7 (cf.~Tab.~\ref{tab:inversion}), round-trip reconstruction values are starting to saturate around $K=50$ to $K=100$ ODE steps. In general, values are lower than on C7, likely because different compounds induce different morphological effects, yielding varying reconstruction difficulty. 

\begin{table}[tb]
\centering
\caption{Inversion fidelity on CK-666 ($N{=}100$ control cells).}
\label{tab:inversion_ck666}
\small
\begin{tabular}{lcc}
\toprule
ODE steps $K$ & PSNR $\uparrow$ & SSIM $\uparrow$ \\
\midrule 
10   & $22.27_{\pm 3.2}$ & $0.815_{\pm 0.06}$ \\
25   & $22.93_{\pm 4.0}$ & $0.830_{\pm 0.06}$ \\
50   & $23.48_{\pm 4.2}$ & $0.838_{\pm 0.06}$ \\
100  & $23.50_{\pm 4.3}$ & $0.839_{\pm 0.06}$ \\
200  & $23.45_{\pm 4.3}$ & $0.839_{\pm 0.06}$ \\
500  & $23.39_{\pm 4.3}$ & $0.838_{\pm 0.06}$ \\
\midrule
VAE only & $38.73_{\pm 3.1}$ & $0.913_{\pm 0.05}$ \\
\bottomrule
\end{tabular}
\end{table}

\paragraph{Distribution Fidelity.}
Table~\ref{tab:ck666_fid_results} reports FID and KID for CK-666, following the same protocol as the C7 evaluation (Sec.~\ref{sec:gen_eval}). Our joint model achieves the best KID at four of five concentrations and the best pooled FID overall, on par with IMPA ($20.85$ vs.\ $20.93$). It surpasses PhenDiff on FID at three of five concentrations and shows less variation in FID across concentrations, indicating more consistent behavior. The ablation (\emph{Ours cond.\ only}) is again uniformly worse, confirming that the benefit of joint $p(z, c)$ modeling is consistent across compounds.

\begin{table}
\centering
\caption{FID/KID comparison (lower is better) for CK-666. %
Best metrics per column are in \textbf{bold}, second-best 
\underline{underlined}.}
\label{tab:ck666_fid_results}
\small
\begin{tabular}{llcccccc}
\toprule
& & \multicolumn{5}{c}{Concentration ($\mu$M)} & \\
\cmidrule(lr){3-7}
Method & Metric & 0.625 & 1.25 & 2.5 & 5.0 & 10.0 & Pooled \\[-0.4em]
 & & \tiny($N{=}1015$) & \tiny($N{=}1033$) & \tiny($N{=}1062$) & \tiny($N{=}1012$) & \tiny($N{=}956$) & \tiny($N{=}5078$) \\
\midrule
Latent CellFlux~\cite{zhang2025cellflux}\textsuperscript{$\dagger$}
    & FID & --- & --- & --- & --- & --- & 21.48 \\
    & KID & --- & --- & --- & --- & --- & 1.60 \\
\midrule
IMPA~\cite{palma2025predicting}\textsuperscript{$\dagger$}
    & FID & --- & --- & --- & --- & --- & \underline{20.93} \\
    & KID & --- & --- & --- & --- & --- & \textbf{1.23} \\
\midrule
PhenDiff~\cite{bourou2024phendiff}
    & FID & \underline{39.88} & \textbf{28.82} & \underline{35.60} & \underline{30.72} & \textbf{31.33} & 22.70 \\
    & KID & \underline{2.58}  & \underline{1.52} & \underline{2.20} & \underline{1.72} & \textbf{1.60} & 1.76  \\
\midrule
Ours (cond.\ only) 
    & FID & 43.59 & 42.43 & 40.31 & 45.21 & 55.70 & 36.28 \\
    & KID & 3.01  & 2.76  & 2.69  & 3.05  & 3.78  & 2.87  \\
\midrule
\textbf{Ours (Joint Latent FM)}
    & FID & \textbf{26.76} & \underline{29.26} & \textbf{27.31} & \textbf{27.84} & \underline{31.99} & \textbf{20.85} \\
    & KID & \textbf{1.39}  & \textbf{1.49}  & \textbf{1.43}  & \textbf{1.49}  & \underline{1.72}  & \underline{1.40}  \\
\bottomrule
\end{tabular}
\vspace{2pt}
\begin{flushleft}
\footnotesize
\textsuperscript{$\dagger$} Latent CellFlux and IMPA do not natively support concentration 
conditioning; therefore we report their pooled results without concentration awareness.
\end{flushleft}
\end{table}

\subsection{Additional Results on UW228}
\label{sec:supp_uw228}
 
\begin{figure}[tb]
    \centering
    \includegraphics[width=\linewidth]{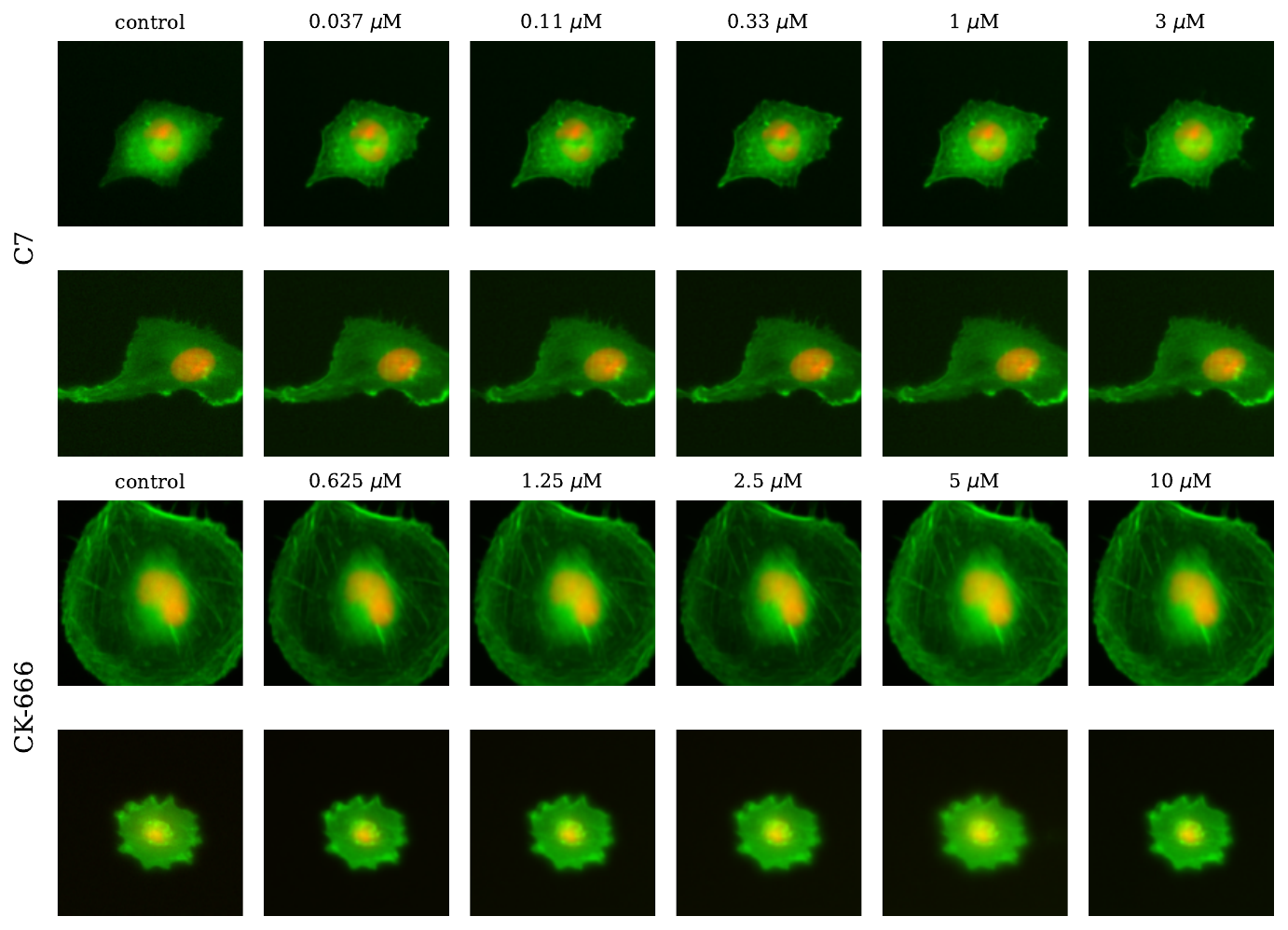}
    \caption{
        \textbf{Qualitative dose-response morphs on the UW228 cell line for 
two compounds (C7 and CK-666), generated by our joint model.} Each row shows a real control cell (leftmost column) morphed to successive concentrations. }
    \label{fig:qualitative_morphs_uw228}
\end{figure}

\new{As a proof of concept for the broader applicability of our approach, 
we apply our joint model to a second cell line (UW228) on the same 
two compounds (C7 and CK-666). Qualitative morphs (Fig.~\ref{fig:qualitative_morphs_uw228}) preserve source-cell identity while producing subtle dose-conditioned variations for both compounds. Quantitatively as shown in Tab.~\ref{tab:uw228_fid}, per-concentration FID and KID stay in a narrow band ($\text{FID}\!\in\![26.3, 30.3]$, $\text{KID}\!\in\![1.17, 1.55]$), and pooled FID is comparable to the ONS76 C7 result reported in the main text 
(Sec.~\ref{sec:gen_eval}) as well as ONS76 CK-666 result reported in~\ref{sec:supp_ck666}, indicating that our method transfers across cell lines without further tuning. A thorough quantitative evaluation with all comparison methods as well as a controllability evaluation remains subject to future work.}
 
\begin{table}[tb]
\centering
\caption{FID and KID for UW228 (our joint model). Metrics are within 
a narrow range per compound and pooled values match those reported 
on ONS76 in the main text.}
\label{tab:uw228_fid}
\small
\begin{tabular}{lcccccc}
\toprule
\multicolumn{7}{c}{C7} \\
\midrule
Concentration ($\mu$M) & 0.037 & 0.11 & 0.33 & 1.0 & 3.0 & Pooled \\
FID & 27.09 & 26.34 & 27.27 & 28.71 & 29.28 & 18.79 \\
KID & 1.32 & 1.24 & 1.34 & 1.26 & 1.36 & 1.25 \\
\midrule
\multicolumn{7}{c}{CK-666} \\
\midrule
Concentration ($\mu$M) & 0.625 & 1.25 & 2.5 & 5.0 & 10.0 & Pooled \\
FID & 26.61 & 27.09 & 27.53 & 30.30 & 27.01 & 19.34 \\
KID & 1.27 & 1.17 & 1.28 & 1.55 & 1.25 & 1.26 \\
\bottomrule
\end{tabular}
\end{table}

\subsection{Extended Concentration Controllability \& Analysis}
\subsubsection{Per-Sample Dose-Response Distributions}

\begin{figure}[tb]
\centering
\includegraphics[width=\linewidth]{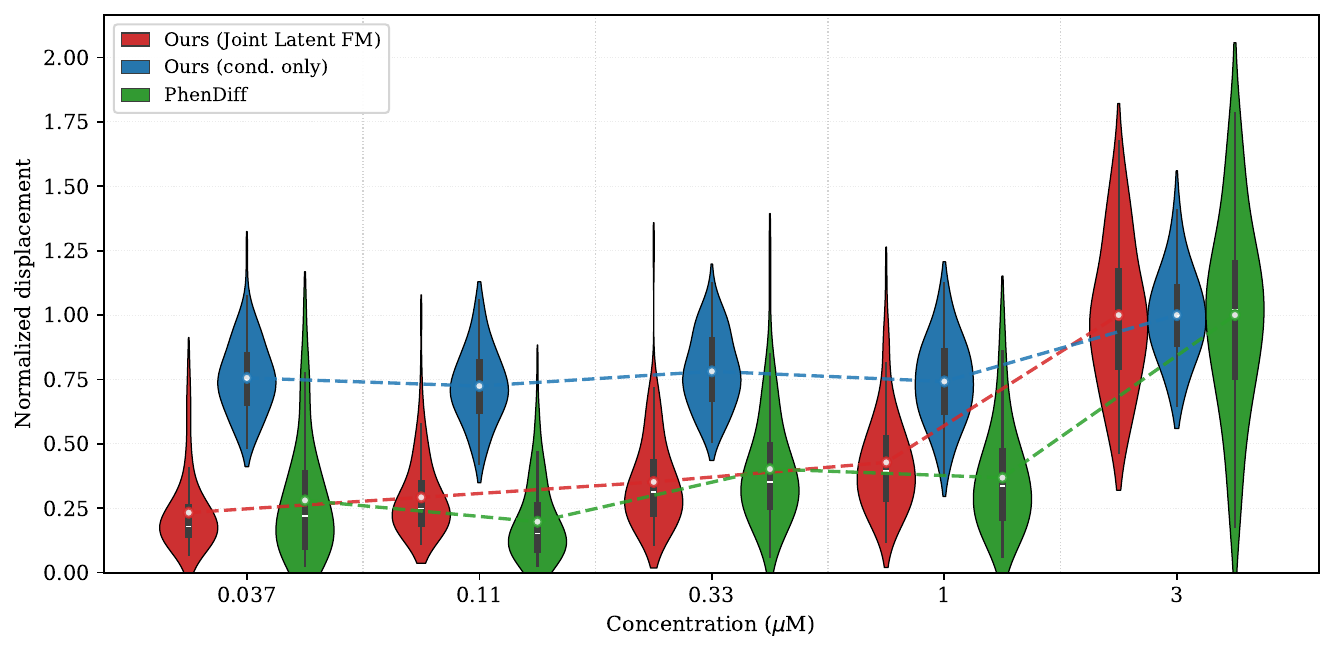}
\caption{\textbf{Per-cell dose-response distributions.}
Violin plots of normalized latent displacement (pixel-space for PhenDiff) from the reference at each concentration; dashed lines connect the concentration means. Normalization per method by the maximum concentration-averaged displacement.} %
\label{fig:violin}
\end{figure}

In Sec.~\ref{sec:controllability}, we studied the latent-space dose response averaged over $n=50$ samples per method, showing the monotonic behavior of our joint model. Figure~\ref{fig:violin} extends this to the per-sample level, showing the distribution of latent displacements underlying Fig.~\ref{fig:trajectory}. The dashed lines mark the concentration-averaged displacement for each method, confirming that our joint model produces the smoothest monotonic mean trajectory across concentrations. In contrast, the conditional-only ablation shows a near-constant mean plateau at low and intermediate doses, only responding at the highest concentration, while PhenDiff exhibits a monotonic mean trend between the two. Per-sample distributions overlap substantially across neighboring doses for all methods, consistent with the intrinsic phenotypic heterogeneity at fixed concentrations.

\subsubsection{Dose-Response Geometry for Unseen Concentrations}
\begin{figure}[tb]
\centering
\includegraphics[width=\linewidth]{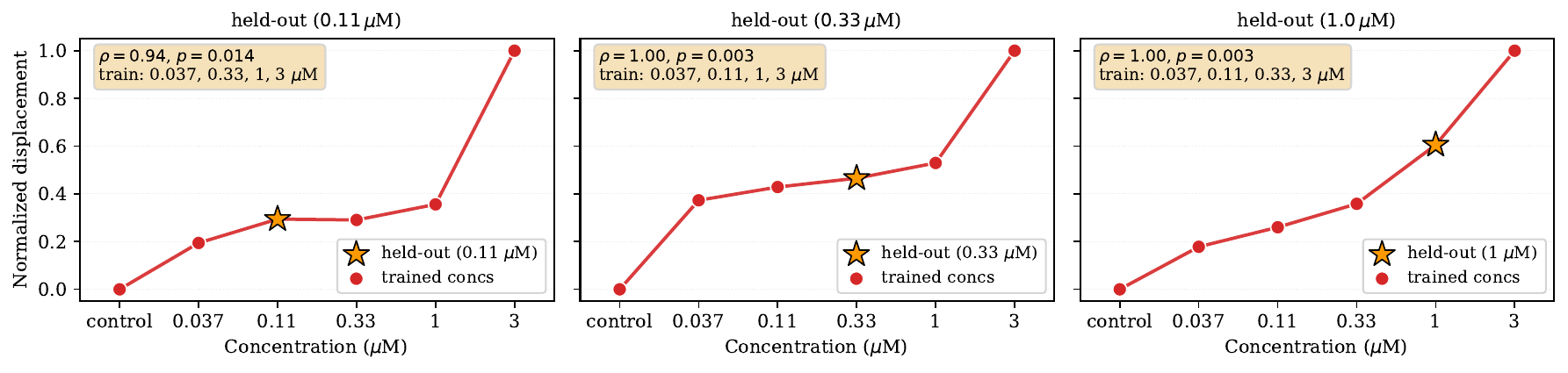}
\caption{
\textbf{Dose-response geometry for unseen concentrations in C7.}
\new{After retraining our joint model with $0.11$, $0.33$, or $1\,\mu\text{M}$ held out during training (panels left to right, respectively), forward integration at the held-out dose (highlighted star) generally interpolates between neighboring concentrations, closely following the expected monotonic trend.}}
\label{fig:holdout}
\end{figure}

\new{To complement the hold-out experiment demonstrated in Sec.~\ref{sec:controllability} with a middle concentration ($0.33~\mu$M) held out on C7, we additionally repeat the same experiment with two additional concentrations held out separately ($0.11~\mu$M and $1~\mu$M). The resulting dose-response geometries can be seen in Fig.~\ref{fig:holdout}, indicating clear monontonic behaviour for $0.33~\mu$M and $1~\mu$M and a monotonic position that slightly exceeds the succeeding one for $0.11~\mu$M, likely due to the fact that lower concentrations are harder to discriminate.}

\subsubsection{Concentration Classification \& Supervised Baseline}
\label{sec:supp_baseline}

\begin{figure}[tb]
\centering
\includegraphics[width=\linewidth]{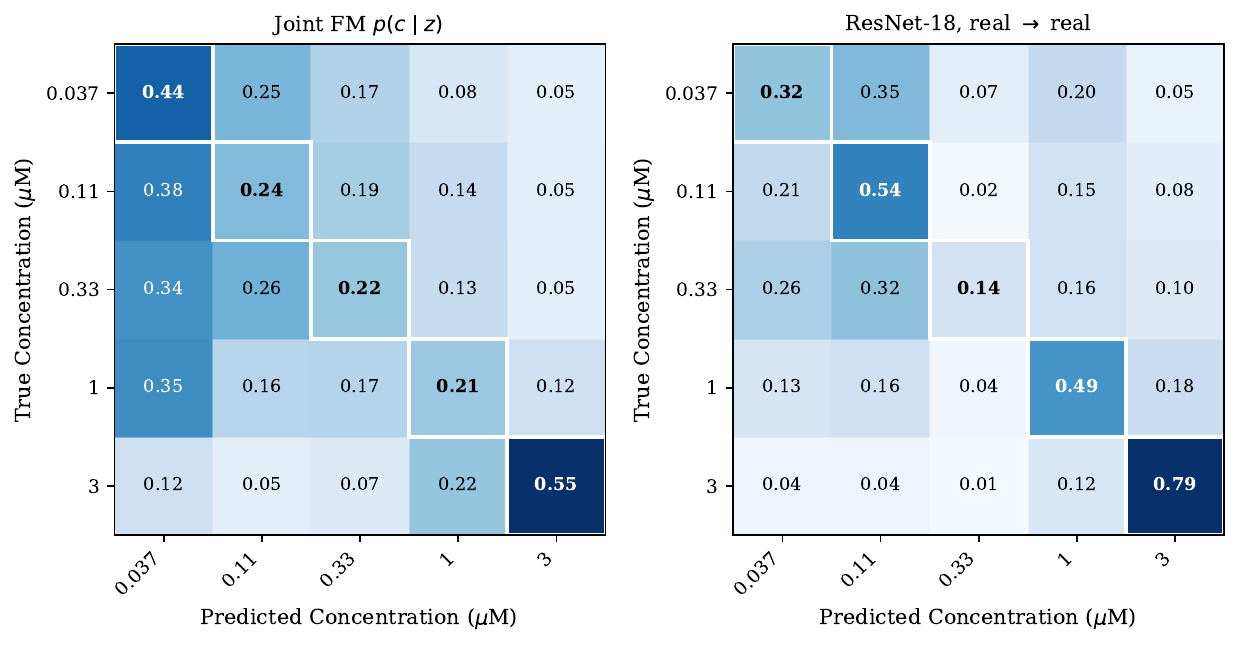}
\caption{
\textbf{Confusion matrices for 5-class concentration prediction in C7.}
\new{\textbf{Left:} Confusion matrix of our joint flow matching model $p(c \mid z)$, showing that lower and intermediate concentrations in particular are harder to discriminate, confirming the findings in Sec.~\ref{sec:controllability}. \textbf{Right:} The corresponding supervised ResNet-18 baseline yields better results at $0.11\,\mu\mathrm{M}$, $1\,\mu\mathrm{M}$, and $3\,\mu\mathrm{M}$, but performs worse at $0.037\,\mu\mathrm{M}$ and $0.33\,\mu\mathrm{M}$.}
}
\label{fig:cms}
\end{figure}

\new{We train a supervised ResNet-18 for multi-class classifications of C7-treated ONS76 cells, providing a supervised ceiling for the concentration estimation task (Sec.~\ref{sec:controllability}). Table~\ref{tab:classifier_comparison} shows the classification accuracy averaged over the 5-class task, whereas Tab.~\ref{tab:perclass_all} and Fig.~\ref{fig:cms} show the per-concentration accuracy as well as the full confusion matrices. 
The supervised ceiling of $48.0\%$ confirms the inherent difficulty of the  task, particularly at intermediate doses where real distributions morphologically overlap (cf.\ Fig.~\ref{fig:conc_matrix}). Our unsupervised $p(c \mid z)$ estimator recovers a substantial share of this ceiling and {outperforms} the supervised baseline at the  lowest concentration ($0.037\,\mu$M: $44.3\%$ vs.\ $31.8\%$), suggesting the continuous concentration formulation enforces  sensitivity to subtle low-dose morphologies.}

\begin{table}[tb]
\centering
\caption{Concentration classification accuracy on C7. 
Best values in \textbf{bold}.}
\label{tab:classifier_comparison}
\small
\begin{tabular}{lcc}
\toprule
Method & Top-1 & $\pm 1$ neighbor \\
\midrule
Joint FM $p(c \mid z)$              & 34.0\% & 68.0\% \\
ResNet-18, real $\to$ real          & \textbf{48.0\%} & \textbf{74.8\%} \\
\bottomrule
\end{tabular}
\end{table}

\begin{table}[tb]
\centering
\caption{Per-concentration accuracy on C7. Best values in \textbf{bold}.}
\label{tab:perclass_all}
\small
\begin{tabular}{lccccc}
\toprule
Concentration ($\mu$M) & 0.037 & 0.11 & 0.33 & 1.0 & 3.0 \\
\midrule
Joint FM $p(c \mid z)$      & \textbf{44.3\%} & 24.1\% & \textbf{21.5\%} & 20.7\% & 54.6\% \\
ResNet-18 (real $\to$ real) & 31.8\% & \textbf{53.9\%} & 14.5\% & \textbf{48.8\%} & \textbf{79.2\%} \\
\bottomrule
\end{tabular}
\end{table}

\subsection{Method Ablations \& Latency Benchmarks}
\subsubsection{Concentration Loss Weight ($\lambda_c$) Sensitivity}
\new{Table~\ref{tab:lambda_ablation} reports FID and KID for three values 
of the concentration-loss weight $\lambda_c \in \{0.1, 0.5, 1.0\}$ on 
C7. Our chosen value ($\lambda_c = 0.5$) achieves the best or 
second-best result at every concentration and metric, confirming 
that the paper's setting is robust and not sensitive to precise 
tuning. Doubling $\lambda_c$ to 1.0 uniformly degrades performance, 
suggesting that over-weighting the concentration flow objective 
harms the primary latent-space generation task; reducing to $0.1$ 
marginally improves fidelity at the lowest concentration but is 
otherwise worse.
Concentration estimation accuracy (Sec.~\ref{sec:controllability}) does 
not strictly correlate with generation quality: $\lambda_c = 0.1$ 
achieves marginally higher top-1 accuracy (36.0\% vs.\ 34.0\%), while 
$\lambda_c = 0.5$ achieves higher $\pm 1$ neighbor accuracy (68.0\% vs.\ 
66.7\%).
}

\begin{table}[tb]
\centering
\caption{Ablation over $\lambda_c$ on C7. 
Best values in \textbf{bold}.}
\label{tab:lambda_ablation}
\small
\begin{tabular}{llcccccc}
\toprule
& & \multicolumn{5}{c}{Concentration ($\mu$M)} & \\
\cmidrule(lr){3-7}
$\lambda_c$ & Metric & 0.037 & 0.11 & 0.33 & 1.0 & 3.0 & Pooled \\
\midrule
\multirow{2}{*}{0.1 \phantom{~}} 
    & FID & \textbf{29.83} & 28.52 & 30.15 & 29.95 & 26.16 & 20.42 \\
    & KID & \textbf{1.56}  & 1.51  & 1.62  & 1.58  & 1.41  & 1.37  \\
\midrule
\multirow{2}{*}{0.5 \phantom{~}} 
    & FID & 30.02 & \textbf{27.20} & \textbf{27.50} & \textbf{27.85} & \textbf{25.02} & \textbf{18.98} \\
    & KID & 1.58  & \textbf{1.40}  & \textbf{1.37}  & \textbf{1.39}  & \textbf{1.28}  & \textbf{1.25} \\
\midrule
\multirow{2}{*}{1.0 \phantom{~}} 
    & FID & 31.48 & 28.21 & 28.91 & 30.41 & 25.66 & 20.32 \\
    & KID & 1.73  & 1.51  & 1.51  & 1.67  & 1.33  & 1.38  \\
\bottomrule
\end{tabular}
\end{table}

\subsubsection{Sampler \& ODE Integration Schemes}
\label{sec:sampler_ablations} 
\new{To complement the inversion fidelity results for C7 in Sec.~\ref{sec:gen_eval}, we perform an ablation on the sampler choice. While all primary experiments use standard Euler integration, we evaluate higher-order alternatives: a second-order Heun solver and a fourth-order Runge-Kutta (RK4) scheme across the same step counts. Table~\ref{tab:sampler_ablation} presents the resulting roundtrip PSNR and SSIM values. At moderate step counts ($K \in \{25, 50\}$), Heun and RK4 yield slight gains in reconstruction quality over Euler; however, performance across all three solvers saturates as step count increases ($K \ge 100$). Given that latent roundtrip fidelity saturates and higher-order solvers yield negligible gains beyond $K \ge 100$ steps, all subsequent forward generation experiments in this work are conducted using standard Euler integration.}

\begin{table}[tb]
\centering
\caption{Inversion fidelity comparison across ODE integration schemes on the C7 test set ($N{=}100$ control cells).}
\label{tab:sampler_ablation}
\small
\setlength{\tabcolsep}{4.5pt}
\begin{tabular}{l ccc ccc}
\toprule
& \multicolumn{3}{c}{PSNR $\uparrow$} & \multicolumn{3}{c}{SSIM $\uparrow$} \\
\cmidrule(lr){2-4} \cmidrule(lr){5-7}
Steps $K$ & Euler & Heun & RK4 & Euler & Heun & RK4 \\
\midrule
10  & $27.79_{\pm 2.9}$ & $27.31_{\pm 2.1}$ & $27.14_{\pm 2.1}$ & $0.866_{\pm 0.04}$ & $0.873_{\pm 0.04}$ & $0.872_{\pm 0.04}$ \\
25  & $31.93_{\pm 3.0}$ & $32.09_{\pm 2.6}$ & $32.00_{\pm 2.6}$ & $0.894_{\pm 0.04}$ & $0.900_{\pm 0.04}$ & $0.900_{\pm 0.04}$ \\
50  & $33.81_{\pm 3.0}$ & $33.99_{\pm 2.8}$ & $33.97_{\pm 2.8}$ & $0.902_{\pm 0.04}$ & $0.906_{\pm 0.05}$ & $0.906_{\pm 0.05}$ \\
100 & $34.50_{\pm 3.0}$ & $34.65_{\pm 2.9}$ & $34.65_{\pm 2.9}$ & $0.905_{\pm 0.05}$ & $0.907_{\pm 0.05}$ & $0.907_{\pm 0.05}$ \\
200 & $34.70_{\pm 3.0}$ & $34.81_{\pm 3.0}$ & $34.81_{\pm 3.0}$ & $0.906_{\pm 0.05}$ & $0.907_{\pm 0.05}$ & $0.907_{\pm 0.05}$ \\
500 & $34.77_{\pm 3.0}$ & $34.81_{\pm 3.0}$ & $34.81_{\pm 3.0}$ & $0.907_{\pm 0.05}$ & $0.907_{\pm 0.05}$ & $0.907_{\pm 0.05}$ \\
\midrule
VAE only & \multicolumn{3}{c}{$38.73_{\pm 3.1}$} & \multicolumn{3}{c}{$0.913_{\pm 0.05}$} \\
\bottomrule
\end{tabular}
\end{table}

\subsubsection{Sampling Times}
\new{Table~\ref{tab:sampling_latency} reports sampling latencies on a single NVIDIA GeForce RTX 4090 (batch size 8, 10 trials) using the default evaluation pipelines from Sec.~\ref{sec:gen_eval}. Our joint model performs VAE encoding, ODE inversion, concentration-conditioned sampling, and VAE decoding, whereas pixel-space baselines (PhenDiff, IMPA) skip VAE transformations. The elevated latencies of our conditional model and Latent CellFlux are due to CFG, which doubles the per-step neural network forward evaluations. While diffusion- and flow-based models split their compute between inversion and forward generation, PhenDiff's main bottleneck is working directly in pixel space rather than a compressed latent space. In contrast, IMPA relies on a single feed-forward pass, making it by far the fastest method.}

\begin{table}[tb]
\centering
\caption{Sampling latency comparison across generative methods and baseline models.}
\label{tab:sampling_latency}
\small
\setlength{\tabcolsep}{5pt}
\begin{tabular}{l c c }
\toprule
Method  & Batch Latency (ms) $\downarrow$ & Per-Cell (ms) $\downarrow$ \\
\midrule
Latent CellFlux   & $4196.5_{\pm 8.2}$  & $524.6$ \\
IMPA              & $176.2_{\pm 0.3}$& $22.0$ \\
PhenDiff          & $26189.5_{\pm 31.9}$ & $3273.7$ \\
Ours (cond. only) & $6367.3_{\pm 51.2}$ & $795.9$ \\
Ours (Joint Latent FM)  & $3237.7_{\pm 31.4}$ & $404.7$ \\
\bottomrule
\end{tabular}
\end{table}

\end{document}